\pdfoutput=1

\documentclass[11pt]{article}

\usepackage[]{acl}

\usepackage{times}
\usepackage{latexsym}
\usepackage{microtype}

\usepackage{graphicx}
\usepackage{booktabs} 
\usepackage{caption}
\usepackage{subcaption}

\usepackage{hyperref}

\usepackage[utf8]{inputenc}
\usepackage{amsmath,amsthm,amssymb,mdframed}
\usepackage{multirow}
\usepackage{xspace}
\usepackage{etoolbox}
\usepackage{stfloats}
\usepackage{latexsym}

\usepackage{tabularx}

\usepackage{graphicx}
\usepackage{enumitem}
\usepackage{floatrow}
\usepackage{bm}
\usepackage{bbm}

\usepackage{blindtext}

\usepackage{wrapfig}
\usepackage{booktabs}
\usepackage{placeins}
\usepackage{xspace}
\usepackage{tabularx}

\usepackage[T1]{fontenc}
\usepackage[T1]{fontenc}

\usepackage[utf8]{inputenc}

\usepackage{microtype}

\usepackage{amsthm,amsmath,amsfonts,bm,xspace}
\usepackage{color}

\newcommand{\comment}[1]{}

\def\eqref#1{(\ref{#1})}

\def\1{\bm{1}}

\def\ve{{\bm{e}}}

\DeclareMathAlphabet{\mathsfit}{\encodingdefault}{\sfdefault}{m}{sl}
\SetMathAlphabet{\mathsfit}{bold}{\encodingdefault}{\sfdefault}{bx}{n}

\title{FACTUAL: A Benchmark for Faithful and Consistent Textual Scene Graph Parsing}

\author{
  Zhuang Li$^1$,Yuyang Chai$^{2,*}$,Terry Yue Zhuo$^{1,3,*}$,Lizhen Qu$^1$, \\
 {\bf Gholamreza Haffari$^1$,Fei Li$^2$,Donghong Ji$^2$,Quan Hung Tran$^4$} \\
  $^1$Monash University,
  $^2$Wuhan University,
  $^3$CSIRO's Data61,
  $^4$Adobe Research \\
  \texttt{\{zhuang.li,terry.zhuo,lizhen.qu,Gholamreza.Haffari\}@monash.edu,} \\
  \texttt{\{yychai,lifei.csnlp,dhji\}@whu.edu.cn,} \\
  \texttt{qtran@adobe.com}
}

\begin{document}
\abovedisplayskip=0.25pt
\abovedisplayshortskip=0.25pt
\belowdisplayskip=0.25pt
\belowdisplayshortskip=0.25pt

\maketitle
\maketitle
\def\thefootnote{*}\footnotetext{The two authors contributed equally to this work.}\def\thefootnote{\arabic{footnote}}
\begin{abstract}


Textual scene graph parsing has become increasingly important in various vision-language applications, including image caption evaluation and image retrieval. However, existing scene graph parsers that convert image captions into scene graphs often suffer from two types of errors. First, the generated scene graphs fail to capture the true semantics of the captions or the corresponding images, resulting in a lack of faithfulness. Second, the generated scene graphs have high inconsistency, with the same semantics represented by different annotations.

To address these challenges, we propose a novel dataset, which involves re-annotating the captions in Visual Genome (VG) using a new intermediate representation called FACTUAL-MR. FACTUAL-MR can be directly converted into faithful and consistent scene graph annotations. Our experimental results clearly demonstrate that the parser trained on our dataset outperforms existing approaches in terms of faithfulness and consistency. This improvement leads to a significant performance boost in both image caption evaluation and zero-shot image retrieval tasks. Furthermore, we introduce a novel metric for measuring scene graph similarity, which, when combined with the improved scene graph parser, achieves state-of-the-art (SOTA) results on multiple benchmark datasets for the aforementioned tasks. The code and dataset are available at \url{https://github.com/zhuang-li/FACTUAL}.

\end{abstract}

\section{Introduction}
A scene graph is a representation that describes the contents of a visual scene, including objects, their attributes, and the relationships between them. The grounding of a scene graph with an image or a text can provide significant benefits for various vision-language tasks, such as image caption evaluation~\cite{anderson2016spice} and image retrieval~\cite{johnson2015image}. Therefore, transduction of image descriptions into scene graphs through textual scene graph parsing has been a crucial vision-language research area.

\begin{figure*}[ht]
    \centering
    \includegraphics[width=0.95\textwidth]{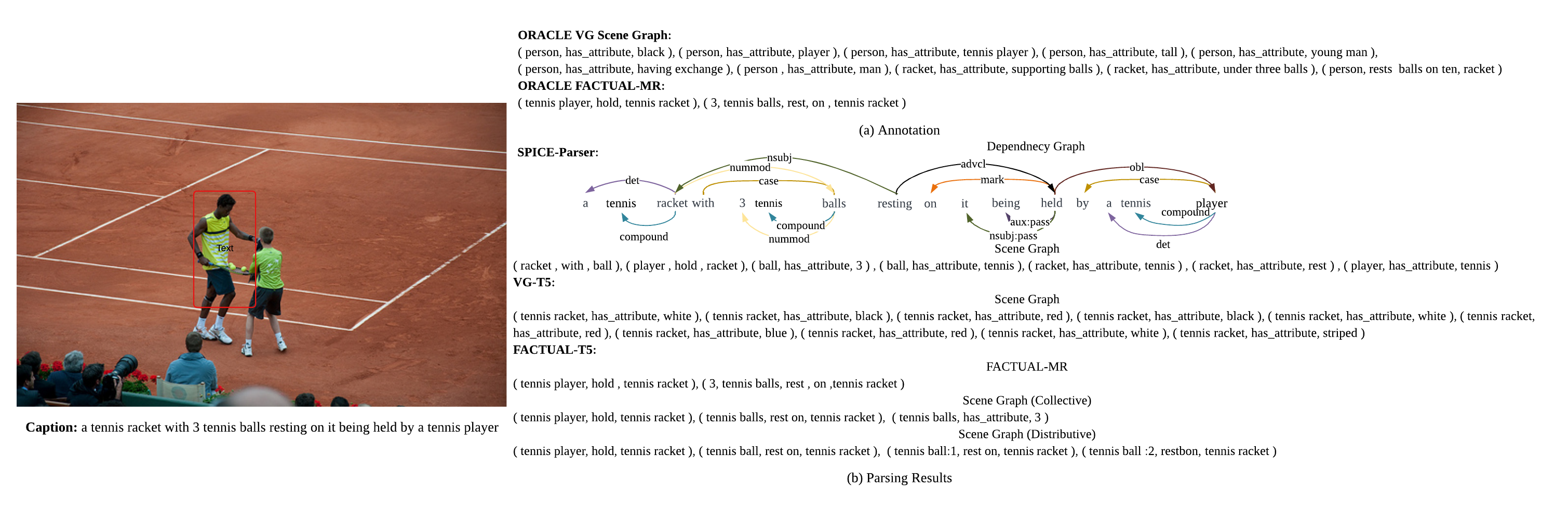}
    \caption{The intermediate representations and scene graphs produced by various parsers are compared with the ORACLE annotations when provided with an image and a caption. \vspace{-3mm}}
    \label{fig:scene_example}
        \vspace{-2mm}
\end{figure*}

Accurately generating scene graphs that capture intersected information from images and their corresponding descriptions is crucial for a successful textual parser. However, current baseline parsers often generate \textit{unfaithful} scene graphs that fail to represent the complete intersected information or generate semantically correct graphs, as shown in Figure~\ref{fig:scene_example}. Furthermore, \textit{inconsistencies} exist in the outputs of scene graph parsers, as depicted in the same figure, where ``tennis'' is interpreted as an attribute in one graph and as a part of an object in another graph. Such inconsistencies can severely impact downstream tasks of textual scene graph parsers, especially when they produce different graphs for a semantic unit, such as a phrase, across various captions, despite they carry the same semantic meaning.

Upon inspection, we hypothesize that the issues of unfaithfulness and inconsistency arise due to the inherent shortcomings of scene graph parsing algorithms and limitations within the datasets. One widely utilized parser, SPICE-Parser~\cite{anderson2016spice}, is known for converting caption dependency graphs into scene graphs using predefined rules, which can result in error propagation. Furthermore, the dependency graphs may not adequately capture the semantic characteristics of scene graphs, as dependency graphs primarily focus on syntactical relationships. Additionally, the limitations of the datasets contribute to the problems as well. As demonstrated in Figure~\ref{fig:scene_example}, the largest scene graph dataset, VG~\cite{krishna2017visual}, includes notable annotation issues regarding faithfulness and inconsistency.


To address the aforementioned issues, we create a high-quality scene graph dataset for training parsers. We firmly believe that the problems of unfaithfulness and inconsistency within the dataset can be effectively resolved by incorporating two key measures: i) employing rigorous definitions for the literals and ii) implementing strict quality control during the annotation process. Therefore, we propose a novel intermediate meaning representation (MR) coined as FACTUAL-MR, which ensures \textbf{FA}ithful and \textbf{C}onsistent tex\textbf{TUAL} scene graph parsing. FACTUAL-MR is a semantic representation that can be deterministically mapped to the scene graph, thereby avoiding the issues that arise from converting syntactical graphs into scene graphs. The annotation of FACTUAL-MRs can be divided into manageable sub-tasks, allowing us to easily control the quality of annotations in each sub-task and ensure their faithfulness. Furthermore, the literals within the FACTUAL-MRs are precisely defined to ensure consistency in textual scene graph parsing annotations. As a result, we re-annotate captions sampled from the VG dataset with FACTUAL-MRs, enabling us to leverage the existing scene graph annotations from VG. Additionally, in order to further enhance the advantages provided by the scene graph parsing for its downstream tasks, we propose a simple yet effective metric called SoftSPICE. This metric calculates graph similarity and significantly improves the performance of vision-language tasks that leverage scene graphs.

Overall, the key contributions are as follows:
\begin{itemize}
\item We propose a novel intermediate representation, FACTUAL-MR, which can be easily annotated and converted into scene graphs. The annotation process of FACTUAL-MR could ensure the faithfulness and consistency of the scene graphs converted from FACTUAL-MR.
\item We construct a large-scale benchmark, FACTUAL, consisting of 40,369 parallel examples. We conduct thorough intrinsic and extrinsic evaluations to demonstrate that FACTUAL significantly improves the performance of textual scene graph parsing.
\item We propose a simple graph similarity metric, SoftSPICE, that achieves new SOTA results in image caption evaluation and zero-shot image retrieval tasks, when combined with a scene graph parser trained with FACTUAL.
\end{itemize}






\section{Related Work}
Grounding a scene graph with an image or image description can be beneficial for a variety of downstream tasks, such as image retrieval~\cite{andrews2019scene,johnson2015image}, image caption evaluation~\cite{anderson2016spice} and image captioning~\cite{zhong2020comprehensive}. Currently, there are three main research directions to scene graph parsing: those that focus on parsing images~\cite{zellers2018neural,tang2020unbiased,xu2017scene,zhang2019graphical,cong2022reltr,li2022sgtr}, text~\cite{anderson2016spice,schuster2015generating,wang2018scene,choi-etal-2022-scene,andrews2019scene,sharifzadeh2022improving}, or both modalities~\cite{zhong2021learning,sharifzadeh2022improving} into scene graphs. Parsing images involves utilizing an object detection model to identify the location and class of objects, as well as classifiers to determine the relationships and attributes of the objects. Textual scene graph parsing employs techniques such as the Sequence-to-Sequence model~\cite{sutskever2014sequence} to parse image descriptions into linearized scene graphs~\cite{sharifzadeh2022improving} or generate intermediate representations, such as dependency graphs or Abstract Meaning Representation (AMR)~\cite{banarescu2013abstract}, which are then mapped into scene graphs using deterministic rules or machine learning models. However, directly utilizing intermediate representations like dependency graphs or AMR often leads to subpar performance in downstream tasks, as emphasized by~\citet{anderson2016spice}, and may even be infeasible for multi-modal tasks requiring annotations for both modalities, given that the intermediate representations only annotate the text. Recent studies in parsing both modalities~\cite{zhong2021learning,sharifzadeh2022improving} have primarily utilized textual parsing models to enhance the performance of visual scene graph parsing. Our work primarily focuses on textual scene graph parsing.

\section{Textual Scene Graph Parsing}
A scene graph, as introduced by~\citet{johnson2015image}, is a formal representation of the objects, their attributes, and the relationships between objects in a visual scene. Given a set of object classes $C$, a set of attribute types $A$, and a set of predicate types $R$, a scene graph $G$ is defined as a tuple $(O, E)$, where $O = \{o_1,...,o_n\}$ is a set of objects and $E \in O \times R \times O$ is the set of edges connecting the objects. Each object $o_i = \{c_i, a_i\}$ is associated with an object class $c_i \in C$ and an attribute $a_i \in A$. 
As depicted in Figure~\ref{fig:scene_example}, our work linearizes a scene graph into a simplified format. In this format, each fact is represented either as $(Object, Has\_attribute, Attribute)$ or as $(Object_{sub}, Predicate, Object_{obj})$, which is consistent with the format of the linearized scene graphs outlined in~\citet{choi-etal-2022-scene,sharifzadeh2022improving}. Therefore, the textual scene parsing aims to learn a mapping $\pi_{\theta}: \mathcal{X} \rightarrow \mathcal{G}$, which translates a textual image description $X \in \mathcal{X}$ into a scene graph $G \in \mathcal{G}$.
\subsection{Challenges}
\paragraph{Unfaithfulness.} 

The scene graph faithfulness is determined by its completeness and correctness.

Completeness is defined as the extent to which the graph conveys the complete semantic meaning of the intersected information from both the caption and the image. For example, Figure~\ref{fig:scene_example} demonstrates that the output of VG-T5~\cite{sharifzadeh2022improving} lacks the facts \textit{(tennis player, hold, tennis racket)} and \textit{(tennis balls, rest on, tennis racket)}, indicating an incomplete graph. This incompleteness issue of parsing outputs can be caused by the noisy training set from VG, which was generated without rigorous quality validation. The other datasets derived from VG also suffer from annotation noise. The customized dependency (CDP) dataset~\cite{wang2018scene} transforms VG scene graphs (VG-SGs) into customized dependency graphs by aligning phrases of objects, attributes, and relations in VG-SGs with corresponding phrases in captions. Consequently, the dependency graphs can be mapped back to scene graphs, referred to as CDP-SGs. Although this approach claims to enhance scene graph parsing performance by framing it as a dependency parsing problem, it results in the loss of additional information due to semantic misalignments between VG-SGs and the captions. As highlighted in Table~\ref{tab:oracle_statistics}, CDP-SGs have more serious completeness issues.

Correctness refers to the semantic accuracy of the graph with respect to the intersected information from the caption and the image. The annotation errors of VG contribute significantly to the correctness issues. As in Figure~\ref{fig:scene_example}, the presence of the predicate ``rest balls on ten'' highlights a significant annotation mistake. Dependency-based parsing methods, such as SPICE-Parser, produce graphs that lack correctness primarily due to error propagation. As shown in Figure~\ref{fig:scene_example}, the term ``rest'' is incorrectly considered an attribute of ``racket'' due to the parsing errors from the Stanford dependency parser~\cite{manning2014stanford}. Another issue with dependency-based methods is that they focus on capturing syntactic relationships among words rather than semantic relationships among objects. The phrases such as ``without leaves'' or ``without a shirt'' indicate the absence of objects like ``leaves'' or ``shirt'' in the scene, but dependency-based methods still interpret them as objects.

\begin{table}[t]
\centering
  \resizebox{0.9\textwidth}{!}{%
  \begin{tabular}{|c|cc|ccc|}
    \toprule
      & \multicolumn{2}{c|}{Faithfulness $\uparrow$} & \multicolumn{3}{c|}{Consistency $\downarrow$}\\
     & Completeness   & Correctness  &  Yules I & TTR & MTLD\\ \hline\midrule 
     VG-SG & 37\% & 29\% & 2.80 & 12.98 &15.17\\
     CDP-SG & 25\% & 73\% & 5.13 &18.69 &27.89\\
     FACTUAL-SG & \textbf{90\%} &  \textbf{78\%} &\textbf{2.37} & \textbf{12.59} & \textbf{15.02}\\
    \bottomrule
  \end{tabular}%
  }
    \caption{ Faithfulness and consistency evaluation of the ORACLE scene graph annotations in VG, CDP, and FACTUAL. We analyze 100 scene graphs extracted from the VG, CDP, and FACTUAL datasets. Our assessment includes measuring the rates of completeness and correctness for these scene graphs. Furthermore, we conduct a comprehensive evaluation of the entire corresponding datasets, utilizing a set of consistency metrics. Please refer to \textbf{Evaluation Metrics} of Sec.~\ref{sec:text_parser} for more details about the evaluation metrics.
     \vspace{-4mm} }
  \label{tab:oracle_statistics}
    \vspace{-3mm}
\end{table}


\paragraph{Inconsistency.} The inconsistency in the dataset is primarily the result of linguistic variations. The object, attribute, and relations are all extracted from texts, but the same semantics can be expressed in multiple ways. For instance, \textit{(tennis player, hold, tennis racket)} and \textit{(tennis racket, held by, tennis player)} are semantically equivalent, even though the orders of the subjects and objects differ. Different understanding of the tasks among crowd workers is also a serious issue. Some may consider ``stone wall'' as a composite object, while others may consider ``stone'' as an attribute and ``wall'' as an object. To measure the consistency of the annotations, we have calculated diversity metrics for the objects, attributes, and predicates within a set of examples encompassing various types of annotations. We assume that the diversity scores indicate the annotations' consistency. As in Table~\ref{tab:oracle_statistics}, the results of the three diversity metrics indicate that the annotations in VG and CDP datasets have a higher degree of diversity regarding their objects, attributes, and predicates than the ones in FACTUAL dataset.

\section{FACTUAL}
\subsection{Meaning Representation}
We propose a novel intermediate \textit{semantic} representation, FACTUAL-MR, in which elements are clearly defined to eliminate confusion among annotators. The task of annotating captions and their associated images with FACTUAL-MRs can be broken down into manageable sub-tasks, and each FACTUAL-MR can be deterministically mapped into a conventional scene graph, enabling the utilization of FACTUAL parser outputs in a wide range of multi-modal applications that rely on scene graphs. Specifically, the template of each fact in FACTUAL-MR is presented in one of two formats: $\{Object, Attribute\}$ or $\{Quantifier_{sub},Object_{sub},Verb,Preposition,\\ Quantifier_{obj},\ Object_{obj}\}$. 
\paragraph{Object.} An object in a scene graph is essentially defined as a grouping of concepts. This results from the widely accepted notion in vision tasks that an image object typically encompasses a collection of homogeneous concepts within a bounding box~\cite{krishna2017visual}. Therefore, a common source of inconsistency in VG-SG is the various methods used to represent the quantity of objects. This can be attributed to the varying understandings of tasks among annotators. For example, as depicted in Figure~\ref{fig:scene_example}, three trees may be represented as a single collective object contained within a large bounding box on an image, with the attribute of ``three'' \textit{(trees, has\_attribute, three)}, or as three distinct objects of \textit{tree} distributed throughout three facts in the visual scene. These different representations of object quantity can lead to inconsistencies. To address this, we propose defining each object in FACTUAL-MR as a grouping of collective concepts. To differentiate between two collective objects with identical names, unique suffix identifiers are utilized. For instance, the phrase ``men watch men'' would be represented as \textit{(men, watch, men:1)}.
\paragraph{Attribute.} The attribute definition in FACTUAL-MR is similar to the original scene graph, with one notable distinction. In FACTUAL-MR, attributes are used to describe all individual concepts within each collective object. For example, in the case of \textit{(3, tennis balls, has\_attribute, white)}, it implies that all the tennis balls are white.
\paragraph{Quantifier.} The quantifier indicates the quantity of concepts within a collective object if the quantity is explicitly mentioned in the text. Additionally, a quantifier modifier may be used to specify the unit of measurement when explicit quantifier modifiers are present in the text. For instance, the phrase ``both men'' is expressed as ``\textit{2, men}'' while ``both groups of men'' would be represented as ``\textit{2g, men}'' and ``both pairs of'' as ``\textit{2p}''. To avoid annotation inconsistencies, a limited set of pre-defined modifiers is provided. In cases where the quantity of objects cannot be expressed by the predefined set, two special quantities, ``\textit{many}'' and ``\textit{unaccountable}'', are offered as placeholders for annotators.
\paragraph{Verb and Preposition.} 
Given the linguistic variations present in VG, the number of relations exceeds 36,000. Through analysis, we have determined that the semantics of each relation can be composed of both a verb and a preposition or either one alone. To this end, we have decomposed these relations into their respective verbs and prepositions. In order to ensure consistency in annotation, a fixed list of verbs and prepositions with exclusive semantics is provided for the annotators to select from. To further facilitate consistency, all verbs are lemmatized to their original forms. The benefits of this decomposition method will be further explained in Section~\ref{sec:annotate}. Additionally, the verb's voice plays a crucial role in the semantics of a fact. For example, the phrases ``cup covered with blanket'' and ``cup covers blanket'' possess distinct semantic meanings. To prevent ambiguity during annotation, an indicator, ``\textit{p:}'', is used as a prefix to the verb to indicate whether it is in a passive voice.
\subsection{Connection to Scene Graph}
\label{sec:connection}
To map a FACTUAL-MR into the original scene graph, we first combine the verb and prepositions into a predicate. The voice of the verb is altered based on whether it is passive or active. However, as the object in our annotation is collective, a collective-distributive ambiguity is present in the sentence, as also highlighted by~\citet{schuster2015generating}. For instance, given an image describing ``three men reading books'', we can know which man is reading which book according to the image, while in the image caption, the information is insufficient to determine this. Previous approaches, such as SPICE~\cite{anderson2016spice} and Stanford~\cite{schuster2015generating} parsers, address this issue using heuristic rules. The SPICE-Parser considers all relations between two collective objects as collective, leading to the phrase being expressed as \textit{(men, reading, books), (men, has\_attribute, 3)}. However, this annotation type is not commonly used as annotators tend to annotate relations distributedly in the VG-SG annotations. Another option, adopted by the Stanford parser, is to consider all these cases as distributive behaviours, resulting in the phrase being expressed as ``\textit{(man, reading, book), (man:1, reading, book), (man:2, reading, book)}''. This may also be incorrect, as three men might read two books. Therefore, in such cases, we improve this heuristic by utilizing our annotated quantifiers. We annotate the implicit quantifiers for the ``books'' according to the image content. If FACTUAL-MR annotates the number of books as three, we know that each man is distributedly reading one book. Otherwise, they are collectively engaging in the activity.


\subsection{Annotation}
\label{sec:annotate}

Our annotation process consists of two stages. In the first stage, we carefully selected approximately 44,000 captions, with each caption aligned to a distinct image, to ensure diversity in our FACTUAL dataset derived from the VG dataset. We hired 25 annotators with diverse backgrounds, either through Amazon Mechanical Turk~\cite{paolacci2010running} or from local undergraduate students, and provided them with one-hour training sessions to ensure consistent annotation practices. Throughout the annotation process, both the images and captions were presented to the annotators to ensure the faithfulness of the annotations to both modalities. Each annotator was reimbursed at a rate of 0.25 USD per task. In the second stage, three expert annotators with a high level of agreement in their annotations performed post-processing and verification steps to ensure the quality of the data. After undergoing the quality check, we retained 40,369 examples in the dataset.


\paragraph{Object and Attribute.} The annotation process for objects and attributes involved extracting information from the captions to ensure faithfulness to the text while utilizing the image to resolve any linguistic ambiguities. For example, in the caption, ``the picture depicts a car'' it is unclear whether the image includes an object labelled as ``picture'' or if the caption is referring to the image itself as a ``picture'' without the context of the image. Furthermore, during the training, the annotators were also instructed to extract the objects for the co-references, such as the pronoun ``it'' mentioned in the captions.
\paragraph{Quantifier.} Regarding quantifiers, the annotators could only select from the pre-determined sets of quantities and quantity modifiers. If an exact match of a modifier was not found, the annotators were instructed to choose the modifier with the equivalent semantic meaning to the modifier in the text. In most cases, only the quantity was annotated when the number of objects was explicitly mentioned. However, exceptions were made for cases involving collective-distributive ambiguity, requiring the annotations of implicit quantities.
\paragraph{Verb and Preposition.} To ensure consistency in the predicate annotations, the annotators were instructed to select from a pre-determined set of predicates rather than writing them on their own. However, the predicates in the VG dataset were not mutually exclusive in semantics. Therefore, we implemented a process of partitioning them into 1000 clusters using K-means, followed by manually selecting around 2000 predicates by observing the clusters. Despite this pruning, the large number of remaining predicates still posed a challenge for annotators to make selections. Therefore, the predicates\footnote{Please note that in some predicates, there are only verbs or only prepositions.} were further decomposed into around 400 verbs and 100 prepositions. For each selection slot, verbs and prepositions were ranked using an information retrieval method, and the annotators were asked to select from the 20 most probable candidates. Annotators were specifically instructed to annotate verbs in the active voice whenever possible. For example, if both active and passive voices were possible for annotation,  as seen in the phrases ``blanket covering cup'' and ``cup covered with a blanket'', both should be annotated as \textit{(blanket, cover, cup)}. However, in cases where only the passive voice construction was syntactically and semantically valid, such as in the example ``cup filled with water,'' it should be annotated as \textit{(cup, p:fill, with, water)} since \textit{(water, fill, cup)} would not be appropriate.
\subsubsection*{Post-processing and Verification.} 
In the second stage, three expert annotators conducted a thorough examination of all cases to verify and rectify annotation errors. Particular attention was paid to identifying and correcting any incorrect annotations related to passive and active voice, as well as quantifiers and their modifiers. Furthermore, in cases where captions did not include specific name phrases for objects but only pronouns, those pronouns were converted into object names. For example, in the sentence ``he is walking'' where ``he'' was annotated as an object, it was resolved to ``man.'' Additionally, any annotations that were entirely unrelated to the text and images were discarded.
\subsection{Statistical Analysis of Dataset}
\begin{table}[t]
\centering
  \resizebox{\textwidth}{!}{%
  \begin{tabular}{|c|c|ccc|c|c|c|}
  \toprule
  & Object & Verb & Prep. & Predicate & Attr. & Quantifier & Fact\\
    \midrule \midrule
     \#Labels & 4,042  & 412& 107& 1,607 & 2,085 &13 & 40,149 \\
     \#Occ. & 116,712 & 25,353 & 32,470 & 70,692 & 22,832 & 2,308 & 71,160\\
     \#Occ. per Class & 28.87 &61.54 &303.46 & 43.99 & 10.95 & 192.33 & 1.77\\\midrule
     \#Labels per Scene & 2.12 & 0.60& 0.78& 1.09 & 0.56 & 0.05& 1.77\\
     \#Occ. per Scene & 2.89 &0.63 & 0.80& 1.75 & 0.57 &0.06 & 1.76\\
    \bottomrule
  \end{tabular}%
  }
    \caption{ The statistics about the number of distinct labels and occurrence (occ.) of the various elements in the 40,369 FACTUAL-MRs. For simplicity, we omit their suffixes when calculating the occurrence of quantifiers.}
  \label{tab:dataset}
    \vspace{-3mm}
\end{table}

We present a statistical overview of the FACTUAL dataset, which comprises 40,369 distinct captions and includes over 4,000 unique object labels with a total occurrence of 116,712. On average, each object label appears approximately 28 times throughout the dataset. Notably, prepositions occur more frequently compared to verbs, although there are four times as many distinct verb labels compared to the number of distinct prepositions. Furthermore, each fact within the dataset tends to be unique within a single caption, with an average occurrence of fewer than two times. Upon analyzing the scene level, we find that, on average, at least two distinct objects are present in each scene. However, there are much fewer distinct verbs, prepositions, and attributes. It is worth highlighting that quantifiers play a relatively minor role in the dataset, as most collective objects described in the image captions consist of only one individual object.

\section{Experiments}
We evaluate the effectiveness of our new scene graph benchmark through one intrinsic evaluation and two extrinsic evaluation tasks.
\subsection{Textual Scene Graph Parsing}
\label{sec:text_parser}
\paragraph{Task Setting.}
Following~\citet{schuster2015generating,wang2018scene,choi-etal-2022-scene}, we construct scene graph parsers to translate textual descriptions of image regions into scene graphs, which are then compared against their respective ground truth scene graphs.
\paragraph{Datasets.} In terms of datasets, our evaluations are conducted on the VG~\cite{krishna2017visual}, CDP~\cite{wang2018scene}, and FACTUAL dataset. The VG dataset comprises 108,077 images and 5.4 million region captions. The CDP dataset converts all scene graphs in VG into a customized dependency graph, which has a one-to-one mapping to the original scene graphs.

We report the performance of the parsers on two data splits for each dataset representation. For the FACTUAL dataset, we consider a random split (Random), which includes 37,861 training, 1,000 validation, and 1,508 test examples. Additionally, we also evaluate a more challenging split (Length) to assess the parsers' compositional generalization abilities. The benchmark test set for this split comprises 1,053 examples. The caption of each example includes more than ten caption tokens and three facts in the corresponding scene graphs. The remaining examples are split into 38,316 training and 1,000 validation examples. The test examples for VG and CDP consist of captions from the Random and Length splits of FACTUAL, while the remaining examples are divided into a validation set of 1,000 and a training set of over 2 million.

\paragraph{Baselines.} In this study, we evaluated the performance of five parsers: \textbf{SPICE-Parser}~\cite{anderson2016spice}, \textbf{AMR-SG-T5}~\cite{choi-etal-2022-scene}, \textbf{CDP-T5}~\cite{choi-etal-2022-scene}, \textbf{VG-T5}~\cite{sharifzadeh2022improving}, and \textbf{FACTUAL-T5}. SPICE utilizes a set of rules to convert dependency graphs of captions into scene graphs. AMR-SG-T5 converts captions into AMRs through the use of AMR-BART~\cite{bai-etal-2022-graph}, and subsequently converts the AMRs into CDP-SG format by using a T5~\cite{raffel2020exploring} model. CDP-T5 directly converts captions into CDP-SGs without the intermediate steps. In contrast to the original CDP-to-SG parser~\cite{wang2018scene}, which relies on intermediate representation, CDP-T5 demonstrates significantly better performance~\cite{choi-etal-2022-scene}. VG-T5, trained on the VG, parses captions into VG-SGs. FACTUAL-T5 parses captions into FACTUAL-SGs and maps them into scene graphs in a collective way. FACTUAL-T5 (pre) was first pre-trained on the VG dataset and then fine-tuned on FACTUAL. As different datasets use different annotations, SPICE\footnote{It is worth noting that SPICE-Parser utilizes a dependency parser trained on a general domain instead of on the CDP dataset. However, it is also based on a dependency parser, and thus we compare its output scene graphs with the ground truth CDP-SGs.}, AMR-SG-T5 and CDP-T5 are evaluated against the ground truth of the CDP dataset, while VG-T5 and FACTUAL-T5 are evaluated against the ground truth VG-SGs and FACTUAL-SGs.

\paragraph{Evaluation.} Following~\citet{schuster2015generating,wang2018scene,choi-etal-2022-scene}, we evaluate scene graph parsers utilizing the SPICE metric~\cite{anderson2016spice}. The SPICE F-score measures the similarity between the candidate and ground truth graph representations extracted from captions by the parsers. In addition, we also employ the Exact Set Match metric~\cite{yu2019sparc}, which assesses the accuracy of the parsers by determining whether the strings of the parsed facts match the ground truth facts while disregarding the order of the facts. During the evaluation, all intermediate representations are converted into scene graphs.

We also evaluate the faithfulness and consistency of parser outputs by human evaluation and automatic lexical diversity metrics, respectively. Specifically, three students manually examine the rates of correctness and completeness of the parsing outputs, and we report the average scores. We employ Yules I~\cite{yule2014statistical}, TTR~\cite{templin1957certain}, and MTLD~\cite{koehn2005europarl} to evaluate the lexical diversity of objects, attributes, and predicates, which indicate consistency of the output scene graphs.
\paragraph{Discussion.}
\begin{table}[t]
\centering
  \resizebox{\textwidth}{!}{%
  \begin{tabular}{|c||cc|cc|}
    \toprule
     \multirow{2}{*}{Parser} &
      \multicolumn{2}{c|}{Random} &
      \multicolumn{2}{c|}{Length}\\
       & Set Match  & SPICE & Set Match & SPICE  \\
      \midrule 
\hline    
   SPICE-Parser & 13.00 & 56.15 & 0.94 & 38.04   \\
   AMR-SG-T5 & 28.45 & 64.82 & 12.16 & 51.71   \\
      CDP-T5 &  46.15 & 73.56 & 26.50 & 61.21  \\
     VG-T5 &  11.54 & 47.46 & 2.94 & 42.98  \\
     \hline    
    FACTUAL-T5 (pre) & \textbf{79.77} & \textbf{92.91} & \textbf{42.35} &  \textbf{82.43}  \\     
           FACTUAL-T5 & 79.44 & 92.23 & 38.65 &  80.76  \\     
    \bottomrule
  \end{tabular}%
  }
    \caption{Intrinsic evaluation results of two metrics for various textual scene graph parsers across two test set splits.
     \vspace{-2mm} }
  \label{tab:parser}
    \vspace{-2mm}
\end{table}
As shown in Table~\ref{tab:parser}, the FACTUAL-T5 and FACTUAL-T5 (pre) models demonstrate a clear superiority over other parsers regarding Set Match and SPICE scores. Notably, the FACTUAL-T5 model, which utilizes the T5 architecture, outperforms other T5-based baselines trained on millions of data points with different annotations. This highlights the effectiveness of the FACTUAL benchmark in generating outputs that are well-aligned with ground truth annotations. In the more challenging Length setting, all parsers experience a decline regarding parsing text into ground truth scene graphs. However, the FACTUAL-T5 model has the least drop among all parsers. Furthermore, pre-training the FACTUAL-T5 model on millions of VG data points only results in a slight improvement in the Length split. This indicates that a dataset as small as 40,000 high-quality examples is sufficient to yield a competent parser.

\begin{table}[t]
\centering
  \resizebox{\textwidth}{!}{%
  \begin{tabular}{|c|cc|ccc|}
    \toprule
      & \multicolumn{2}{c|}{Faithfulness $\uparrow$} & \multicolumn{3}{c|}{Consistency $\downarrow$}\\
     & Completeness   & Correctness  &  Yules I & TTR & MTLD\\ \hline\midrule 
     SPICE-Parser & 49\% & 57\% & 1.56 & 10.26 &14.87\\
     AMR-SG-T5 & 31\% & 71\% & 2.85 & 15.45 &22.56\\
     CDP-T5 & 28\% & 86\% & 3.64 &16.57 &23.96\\
     VG-T5 & 51\% & 47\% & \textbf{0.37} &\textbf{5.27} &\textbf{10.59}\\
     \textbf{FACTUAL-T5 (pre)} & \textbf{92\%} &  \textbf{93\%} &2.76 & 13.55 & 15.30\\
    \bottomrule
  \end{tabular}%
  }
    \caption{Evaluation of faithfulness and consistency across outputs from various scene graph parsers. }
  \label{tab:parser_faith_consistency}
    \vspace{-3mm}
\end{table}
The SPICE-Parser has become the most frequently utilized parser in vision-language tasks. However, as shown in Table~\ref{tab:parser}, it is unable to align with the CDP-SG in either of the two settings. However, this does not necessarily imply that the SPICE-Parser is the worst among the parsers, as the oracle CDP-SGs have a high degree of noise as well, as demonstrated in Table~\ref{tab:oracle_statistics}. Our human evaluation of the faithfulness of the parsing results, as presented in Table~\ref{tab:parser_faith_consistency}, indicates that the SPICE-Parser can perform comparably with the VG-T5 model and outperform the CDP-T5 model in terms of completeness. Furthermore, our subsequent extrinsic evaluation also shows that the SPICE-Parser is the second-best parser among the parsers evaluated. Table~\ref{tab:parser_faith_consistency} also illustrates that our parser performs much better than the other baselines in terms of faithfulness while ranking second in terms of consistency. Interestingly, the VG-T5 model exhibits the best performance in consistency. However, its ORACLE annotations are more inconsistent than ours. Our analysis reveals that the VG-T5 prioritizes predicting scene graphs with simple lexicons and discards more complex patterns, resulting in its strong performance in consistency but much weaker performance in faithfulness metrics.

\subsection{Image Caption Evaluation}
\begin{table}[t]
\centering
  \resizebox{\textwidth}{!}{%
  \begin{tabular}{|cc||cc|cc|}
    \toprule
    \multirow{2}{*}{Metric}& \multirow{2}{*}{Parser} &
      \multicolumn{2}{c|}{Flicker8K} &
      FOIL (1-ref)& FOIL (4-ref)\\
      & & $\tau_c\uparrow$  & $\rho\uparrow$ & $Acc\uparrow$  & $Acc\uparrow$  \\
      \midrule  \midrule
   \multirow{3}{*}{SPICE} & SPICE-Parser & 44.77 & 60.11 & 76.31 & \textbf{87.02}  \\
        & CDP-T5 & 33.50 & 49.50 & 65.66 & 72.76  \\
       & VG-T5 & 37.18 & 51.94 & 68.43 & 76.12 \\

    & \textbf{FACTUAL-T5(pre)}  & \textbf{45.12} & \textbf{60.78} & \textbf{76.69} & 86.88  \\
\hline    
 \multirow{3}{*}{SoftSPICE} &  SPICE-Parser & 51.897 & 68.118 & 78.53 & 86.77   \\
     & CDP-T5 & 45.54 & 59.64 &53.58 & 59.49   \\
    & VG-T5 & 39.66 & 53.05 &70.80 & 76.77   \\
   & \textbf{FACTUAL-T5(pre)} & \textbf{53.35} & \textbf{69.52} & \textbf{85.66} & \textbf{91.61}   \\     
    \bottomrule
  \end{tabular}%
  }
    \caption{ (Left) The correlation scores between SPICE or SoftSPICE with the human judgment. (Right) The accuracies of the metrics w.r.t. detecting the hallucinated sentences.
     \vspace{-3mm} }
  \label{tab:img_eval}
    \vspace{-3mm}
\end{table}

\paragraph{Task Setting.} To assess the quality of the model-generated captions regarding a set of reference captions and an image, we adopt the SPICE and SoftSPICE metrics to calculate a graph similarity between graphs extracted from the candidate and reference captions. As these metrics are based on the parser outputs, a \textit{better} parser will result in scores that more closely align with human judgment. 
\paragraph{Evaluation.} Following~\citet{hessel2021clipscore}, we employ two evaluation settings. The first setting involves calculating the correlation of the scores with human judgment utilizing Kendall's $\tau$ and Pearson correlation on the Flicker8K dataset~\cite{hodosh2013framing}. The Flicker8K dataset includes 17k "expert" human judgments for 5664 images, with each caption being rated on a scale of 1 to 4 against five reference captions. In the second setting, we utilize one (1-ref) or four (4-ref) reference captions sourced from the FOIL dataset~\cite{shekhar2017foil}. This dataset consists of 32k pairs of true captions and their corresponding corrupted versions, where a single word is replaced with an incorrect one. The objective is to assess the accuracy of each image caption evaluation metric in identifying and assigning higher scores to the uncorrupted captions. This setting aims to evaluate the metric's ability to detect instances of sentence hallucination effectively.

\paragraph{SoftSPICE.} SPICE calculates the similarity between two graphs by matching strings of sub-components within the graphs. These sub-components include \textit{objects}, tuples \textit{\{object, attribute\}} and triples \textit{\{object, predicate, object\}}. To improve SPICE, we propose an alternative method that utilizes embedding-based techniques to calculate string similarity. This approach involves decomposing each graph into the aforementioned sub-components and encoding the text of each component using the Sentence-BERT~\cite{reimers2019sentence}. The resulting similarity score, coined SoftSPICE, is as follows:

\begin{small}
\begin{align}
    \phi_s(G_c,G_r) = \frac{1}{|\mathcal{V}_c|}\sum_{\ve_c\in\mathcal{V}_c}\max_{\ve_r\in\mathcal{V}_{r}}(cos(\ve_c,\ve_r))
\end{align}
\end{small}
\noindent where $\ve$ denotes the embedding of each component, $\mathcal{V}_{r}$ and $\mathcal{V}_{c}$ denote the sets of embeddings encoding components within the candidate and reference graphs, respectively. Additionally, we can also use the image $I$ to compute a \textbf{SoftSPICE(img)} score, denoted as $\phi_i(G_c,I)$. This score is computed by combining the embeddings of the graph components and the image:

\vspace{-2mm}
\begin{small}
\begin{align}
    \phi'_i(G_c,I) &= \frac{1}{|\mathcal{V}_c|}\sum_{\ve_c\in\mathcal{V}_c}cos(\ve_c,\ve_I)\\
    \phi_i(G_c,I) &= \frac{2\cdot\phi_s(G_c,I)\cdot\phi'_i(G_c,I)}{\phi_s(G_c,G_r)+\phi'_i(G_c,G_r)}
\end{align}
\end{small}

\noindent
where $e_c$ and $e_I$ are obtained by encoding the sub-components and the images with CLIP.

\paragraph{Discussion.} Table~\ref{tab:img_eval} illustrates that FACTUAL-T5 demonstrates improvement over other parsers in terms of enhancing the correlation of SPICE and SoftSPICE scores with human judgments. However, when using SPICE to detect hallucinated instances, our parser performs comparably to the SPICE-Parser. We attribute this to the fact that approximately one-third of the pairs will have tied SPICE scores due to the use of exact string matching. On the other hand, when using the embedding-based metric, SoftSPICE, the superiority of our parser on FOIL is revealed. Currently, the SPICE utilizing the SPICE-Parser has been a common standard in image caption evaluation settings. We are confident that our parser can be a suitable replacement for SPICE-Parser. 
\begin{table}[t]
\centering
  \resizebox{\textwidth}{!}{%
  \begin{tabular}{|c||cc|cc|}
    \toprule
    \multirow{2}{*}{Metric} &
      \multicolumn{2}{c|}{Flicker8K} &
      FOIL (1-ref)& FOIL (4-ref)\\
       & $\tau_c\uparrow$  & $\rho\uparrow$ & $Acc\uparrow$  & $Acc\uparrow$  \\
      \midrule  \midrule
 SoftSPICE & 53.35 & 69.52 & 85.66 & 91.61  \\
         SoftSPICE(img) & 54.85 & 70.55 & 88.12 & 92.31   \\
         \hline
         BERTScore & 36.71 & 49.81 & 86.70  & 90.49  \\
      BERTScore + SoftSPICE(img)& 51.08 & 65.80 & 90.50  &  \textbf{94.64} \\
          \hline
        CLIPScore & 51.44 & 64.86 & 86.85  & 86.85  \\
        RefCLIPScore & 53.00 & 67.67 & \textbf{90.94}  & 92.40  \\
                RefCLIPScore + SoftSPICE(img) & \textbf{57.37} & \textbf{73.40} & 90.69  &94.01\\

    \bottomrule
  \end{tabular}%
  }
    \caption{ The results comparing SoftSPICE with current SOTA image caption evaluation metrics. We use FACTUAL-T5 as the parser for SoftSPICE. 
     \vspace{-2mm} }
  \label{tab:img_eval_sota}
    \vspace{-3mm}
\end{table}

We also compare SoftSPICE with current SOTA image evaluation metrics, namely BERTScore~\cite{zhang2019bertscore}, CLIPScore, and RefCLIPScore. These metrics calculate the similarity between the embeddings of the candidate caption with the embeddings of the reference captions, the image, and both reference captions and images, respectively. As in Table~\ref{tab:img_eval_sota}, SoftSPICE performs comparably with all the SOTA methods \textit{when there are over four reference captions}, and with the inclusion of image information, SoftSPICE(img) can even outperform SOTA results on Flicker8K. We also observed that the scene graph feature could be a useful supplement to caption-level features. By taking the harmonic mean of SoftSPICE(img) with BERTScore and RefCLIPScore, the performance of both metrics achieve new SOTA results.
\subsection{Zero-shot Image Retrieval}
\paragraph{Task Setting.} The goal of image retrieval is to identify and retrieve an image that precisely corresponds to a given textual query description. This is typically accomplished by allocating scores to images based on their relevance to the query and selecting the top $k$ images. 

Following the setting from~\citet{johnson2015image,wang2018scene}, we have selected 456 captions and their corresponding images from the Random and Length test sets, initially prepared for intrinsic evaluation. These captions serve as queries to retrieve their associated images, forming the basis for evaluating the performance of our image retrieval system. We proceed under the assumption that an oracle scene graph corresponding to each selected image is available. Furthermore, we introduce a '\textit{Local}' setting, which provides access to the coordinates of a bounding box within each image that corresponds to each caption and the ground truth scene graph aligned with this bounding box region. 

\paragraph{Evaluation.} During the evaluation, the scene graph of the captions is generated using various baseline parsing methods. The 456 images are ranked according to the similarity scores computed using either the SoftSPICE or CLIPScore between each image and the caption. Notably, the representation encoders employed in both similarity measurements are not fine-tuned on the in-domain dataset. The performance of various methods is assessed using the Recall@k metric. The performance of different methods is assessed using the Recall@k metric, which indicates the percentage of caption queries where the top $k$ retrieved images, given a specific query, include the ground truth.

\paragraph{Discussion.} 
As observed in Table~\ref{tab:img_retrieval}, FACTUAL-T5 consistently outperforms other baselines in zero-shot image retrieval tasks, highlighting the superiority of our dataset and parser. The performance of both SoftSPICE and CLIPScore is generally enhanced by incorporating location information of the bounding boxes, depicting that more accurate information could boost image retrieval. Moreover, when combined with all available parsers, SoftSPICE demonstrates significantly superior performance compared to CLIPScore, emphasizing the substantial potential benefits of utilizing structured information for image retrieval. 
\begin{table}[t]
\centering
  \resizebox{\textwidth}{!}{%
  \begin{tabular}{|ccc||cc|cc|}
    \toprule
    & \multirow{2}{*}{Method} & \multirow{2}{*}{Parser} &
      \multicolumn{2}{c|}{Random} &
      \multicolumn{2}{c|}{Length}\\
      & & & R@1  & R@5 & R@1 & R@5  \\
      \midrule  \midrule
   \multirow{4}{*}{Local.} &\multirow{3}{*}{SoftSPICE} &  SPICE-Parser & 67.76 & 84.87 & 67.54 & 81.80  \\
   & &  CDP-T5 & 72.59 & 88.16 & 62.28 & 80.70  \\
   & & VG-T5 &49.56  & 68.86 & 58.77 & 74.34 \\

&    & FACTUAL-T5  & \textbf{79.39} & \textbf{92.32} & \textbf{75} & \textbf{87.06}  \\

    &\multirow{1}{*}{CLIPScore} & N/A  & 31.58 & 58.77 & 45.61 & 66.01  \\
        \hline
\hline    
 \multirow{4}{*}{No Local.} &\multirow{4}{*}{SoftSPICE} &   SPICE-Parser & 47.81 & 71.05  & 57.01 & 78.07   \\
 &&   CDP-T5 & 57.02 & 76.54 & 51.54 & 71.27   \\
    && VG-T5 & 38.38 & 58.11 &51.54 & 70.61   \\
   && FACTUAL-T5 & \textbf{66.45} & \textbf{83.99} & \textbf{68.42} & \textbf{85.53}   \\   
      
       &\multirow{1}{*}{CLIPScore} & N/A  & 23.02 & 47.37 & 34.65 & 55.26  \\
    \bottomrule
  \end{tabular}%
  }
    \caption{ Zero-shot image retrieval evaluation on two sets of image-caption pairs that utilize localization or do not use localization information during image retrieval.
     \vspace{-3mm} }
  \label{tab:img_retrieval}
    \vspace{-3mm}
\end{table}

\section{Conclusion}
We introduce a new intermediate representation, coined FACTUAL-MR, which aims to address the issues of faithfulness and consistency for textual scene graph parsers. By utilizing a rigorous annotation process, it is possible to create a large-scale dataset based on FACTUAL-MR. Our experiments demonstrate that FACTUAL-T5, trained on this dataset, is capable of generating consistent scene graphs that are highly faithful to corresponding images and captions. Utilizing a novel graph similarity metric, SoftSPICE, FACTUAL-T5 significantly improve performance in both image caption evaluation and zero-shot image retrieval.

\section{Limitations}
Despite the significant advancements made by the proposed FACTUAL-MR representation in addressing the limitations of current scene graph parsing datasets, there remain several areas for future research.

First, FACTUAL-MR currently relies on heuristic rules to resolve the collective-distributive ambiguity as introduced in Section~\ref{sec:connection}. However, the limitations still remain due to the ambiguity of language. To obtain a perfect parser, rich-world knowledge from multi-modalities or textual context~\cite{li2020context} is required, which is left as our future work.

Second, there is currently no explicit alignment between objects represented within FACTUAL-MR and the corresponding bounding boxes in the image. To fully utilize multi-modal information, collecting such alignments may be necessary.

Third, the proposed method utilizes ORACLE scene graphs of the image, however, in practical applications, extracting a scene graph from an image remains a challenging problem. Further research is required to determine if utilizing a visual scene graph parsing model to extract scene graphs from images would negatively impact image retrieval performance.

Lastly, our current approach utilizes a large pre-trained language model to train the parser. However, the issue of robustness in parsers~\cite{huang2021robustness,zhuo2023robustness} has always been a significant concern. The captions in the VG dataset mainly consist of short sentences with simple patterns. It remains unclear whether the parser is robust enough to handle sentences with more complex linguistic variations, which calls for further investigation.

\section*{Acknowledgments}
We would like to express our gratitude to Weibo Shi for his valuable assistance in conducting our human evaluation works. We also extend our appreciation to Adobe Inc. for their generous funding support in data collection. Additionally, we would like to thank Wuhan University for their valuable assistance in identifying students to assist with data annotation.

\bibliography{anthology,custom}
\bibliographystyle{acl_natbib}

\end{document}